\documentclass{article}

\usepackage{arxiv}

\usepackage[utf8]{inputenc} % allow utf-8 input
\usepackage[T1]{fontenc}    % use 8-bit T1 fonts
\usepackage{hyperref}       % hyperlinks
\usepackage{booktabs}       % professional-quality tables
\usepackage{amsfonts}       % blackboard math symbols
\usepackage{nicefrac}       % compact symbols for 1/2, etc.
\usepackage{microtype}      % microtypography
\usepackage{lipsum}
\usepackage{authblk}
\usepackage{graphicx}
\usepackage{amsmath}
\usepackage{comment}
\usepackage{xcolor}
\usepackage{adjustbox}
\usepackage{booktabs}
\usepackage{subcaption}
\usepackage{float}
\usepackage{algorithmic}
\usepackage{caption}
\usepackage{todonotes}

\definecolor{codegreen}{rgb}{0,0.6,0}
\definecolor{codegray}{rgb}{0.5,0.5,0.5}
\definecolor{codepurple}{rgb}{0.58,0,0.82}
\definecolor{backcolour}{rgb}{0.95,0.95,0.92}

\usepackage{listings}
\lstdefinestyle{mystyle}{
    backgroundcolor=\color{white},   
    commentstyle=\color{codegreen},
    keywordstyle=\color{magenta},
    numberstyle=\tiny\color{codegray},
    stringstyle=\color{codepurple},
    basicstyle=\ttfamily\footnotesize,
    breakatwhitespace=false,         
    breaklines=true,                 
    captionpos=b,                    
    keepspaces=true,                 
    numbers=left,                    
    numbersep=5pt,                  
    showspaces=false,                
    showstringspaces=false,
    showtabs=false,                  
    tabsize=2
}
\title{\textit{KnowIt}: Deep time series modeling and interpretation}

\author[1,2,3]{\textbf{M.~W.~Theunissen}}
\author[1, 2]{\textbf{R.~Rabe}}
\author[1, 2]{\textbf{H.~L.~Potgieter}}
\author[1,2,3]{\textbf{M.~H.~Davel}}

\affil[1]{Faculty of Engineering, North-West University, South Africa}
\affil[2]{Centre for Artificial Intelligence Research (CAIR), South Africa}
\affil[3]{National Institute for Theoretical and Computational Sciences (NITheCS), South Africa}

\begin{document}

\maketitle

\begin{abstract}
\textit{KnowIt} (\textbf{Know}ledge discovery \textbf{I}n \textbf{t}ime series data) is a flexible framework for building deep time series models and interpreting them.
It is implemented as a Python toolkit, with
source code and documentation available from \url{https://must-deep-learning.github.io/KnowIt}. It imposes minimal assumptions about task specifications and decouples the definition of dataset, deep neural network architecture, and interpretability technique through well defined interfaces. This ensures the ease of importing new datasets, custom architectures, and the definition of different interpretability paradigms while maintaining on-the-fly modeling and interpretation of different aspects of a user's own time series data. \textit{KnowIt} aims to provide an environment where users can perform knowledge discovery on their own complex time series data through building powerful deep learning models and explaining their behavior.
With ongoing development, collaboration and application our goal is to make this a platform to progress this underexplored field and produce a trusted tool for deep time series modeling.
\end{abstract}

% keywords can be removed

\section{Introduction}

Many scientific and engineering applications rely on data collected chronologically. 
Such data can display highly complex time-varying interactions among observed variables. 
This has led to the increased use of deep neural networks (DNNs) as a modeling choice, given their performance on complex, data-driven tasks~\cite{wang2024tssurvey}.
%, as demonstrated in domains as diverse as finance, medicine, ecology and industry.  
While potentially accurate, DNNs are not intrinsically interpretable: add-on techniques are required to describe the reasoning behind a model’s decision in a way that would be understandable to a human~\cite{rojat2021explainable}. 

The interpretability of a model is particularly important in three cases: 
it can be a requirement in high-risk applications; 
it can be necessary to ensure fairness and lack of unwanted model bias; and 
interpretable models can be used to uncover unexpected patterns in data, contributing to a better understanding of the underlying tasks. 
These all require not just the development of accurate time series models, but the ability to extract explanations, either of individual decisions or of the model as a whole. 
Where applications are multivariate or utilize data across multiple time steps, explanations themselves can quickly become uninterpretable, requiring specialized tools to ``explain the explanations''~\cite{schlegel2019towards, theissler2022explainable}. 

The {\it KnowIt} toolkit aims to facilitate knowledge discovery in time series data by providing a DNN-based experimental platform for modeling and interpreting time series data.
It is novel in providing:
%Motivate need for interpretable deep time series models. Set the scene. Novelty:
%\begin{itemize}
    %\item 
    (i) a flexible experimental setup for interpretability and knowledge discovery;
    %\item 
    (ii) on-the-fly modeling of different aspects of the same data;
    %\item 
    (iii) a focus on deep time series modeling, specifically; and
    %\item 
    easy importing of new datasets and architectures.
%\end{itemize}
%\todo{Add paper intro sentence once structure finalised - see Section 2 comment}

The toolkit implements a flexible framework for time series modeling, as described in the next section.
We provide an overview of the code design (Section \ref{sec: code design}) and toolkit usage (Section \ref{sec: operattional flow}), before contrasting this implementation with related toolkits in Section \ref{sec: related toolkits}.

\section{Time series modeling}

% Focus of toolkit: time series tasks, architectures, interpretability techniques.

% As stated previously, the goal of \textit{KnowIt} is to provide the user with a flexible environment to build powerful deep time-series models and interpret them. 

At the highest level, \textit{KnowIt} builds and analyzes models of one or more features varying over time. 
These features may require classification, regression to a target value, autoregression of observed features, or related tasks.

In order to dynamically facilitate appropriate splitting, scaling, sampling and padding for DNN training, it is expected that each time step is equidistant from the preceding and following time step, separated by some time delta. Furthermore, \textit{KnowIt} expects that the timesteps are timestamped. 
% using a pandas
% \textit{DatetimeIndex}~\cite{The_pandas_development_team_pandas-dev_pandas_Pandas}. 
If the data lacks entries for some timesteps, the data will be split into ``slices'', which represent internally contiguous blocks of time. Missing timesteps can also be filled by means of interpolation.

Time series modeling is a broad topic. %It is worth narrowing down the type of modeling performed within \textit{KnowIt}. 
In Section~\ref{sec: problem class} we define the specific problem class that \textit{KnowIt} addresses, in Section~\ref{sec: archs} we describe the considered deep learning architectures, and in Section~\ref{sec: interpret} we mention the currently supported interpretability methods.

\subsection{Problem class}
\label{sec: problem class}

\textit{KnowIt} supports two general problem classes, as defined below.

\subsubsection{Regression and Classification}
Deep time series modeling, in the \textit{KnowIt} environment, is defined with respect to \textit{prediction points}. These are exact points in time, relative to which the input and output features of the model are defined through time delays. 
As a simple example: a model might predict the temperature in six hours based on the temperature and wind-speed, measured each hour, for the last three hours. In this scenario, the model uses 
\begin{itemize}
    \item input time delays = $[-2, 0]$,
    \item input components = \textit{temperature} and \textit{wind-speed},
    \item output time delays = $[6, 6]$\footnote{Meaning from six timesteps into the future to six timesteps into the future.}, and 
    \item output component = \textit{temperature}.
\end{itemize}

We generalize this framework so that the input and output time delays and components of the model can be arbitrarily defined relative to the prediction point. This allows us to define various tasks by dictating characteristics such as whether the prediction is autoregressive (output components are also in the input), multi-step (multiple timesteps contained in the output), or strictly causal (the inputs and outputs do not overlap in time and the input precedes the output). For a more detailed definition of this setup, see Appendix\ref{app sec: problem class}.

Note that, at first glance, this problem class seems to preclude any task that requires the processing of variable-length blocks of time, but this is not the case. 
Internally, the models have fixed-length input windows. Variable-length inputs are handled through techniques like padding, truncation, sliding windows, aggregation, and down- or up-sampling. We have chosen to define the problem class with respect to the internal fixed-length window to provide the user with more flexibility and control over what is being modeled and interpreted. 
% We aim to introduce the ability to build stateful models in the KnowIt environment, which will allow the processing of variable-length blocks of time while maintaining the problem definition described above.

If a task requires that the model consider information beyond its input window, \textit{KnowIt} also facilitates stateful training, where a hidden state is maintained across prediction points, often packaged into chronologically ordered batches. This allows the model to learn patterns across variable-length blocks of time while still maintaining the problem class as defined above. See the \textit{Statefulness tutorial}~\footnote{\url{https://must-deep-learning.github.io/KnowIt/markdowns/tutorials/stateful/stateful_tut_readme.html}} for an illustrative example.

\subsubsection{Variable length regression}

For use cases where it is easier (or necessary) to define the model as producing variable length predictions based on variable length inputs \textit{KnowIt} also supports \textit{variable length regression}. In this problem class the output will always contain as many time steps as the input, possibly with some defined delay. A corresponding example to the previous one would be if the model predicts the temperature in six hours based on the temperature and wind-speed, measured each hour, in the past. In this scenario, the model uses
\begin{itemize}
    \item input time delays = $[0, 0]$,
    \item input components = \textit{temperature} and \textit{wind-speed},
    \item output time delays = $[6, 6]$, and 
    \item output component = \textit{temperature}.
\end{itemize}

Note that the time delays no longer define a fixed window, but simply the delay with which the corresponding components are presented to the model. Also note that classification is not currently supported for this problem class.

\subsection{Architectures}
\label{sec: archs}

All architectures in the \textit{KnowIt} environment use the same task-specific input and output shapes. Whether, and in what way, the resulting model considers the temporal nature of the data across the input time delays is determined by the choice and construction of the architecture that will be used to build the model.

There is a set of default architectures that the user can choose from, or the user can define a custom architecture which meets the conditions defined in the \textit{Architecture How-to}~\footnote{\url{https://must-deep-learning.github.io/KnowIt/markdowns/guides/archs_readme.html}} documentation. Currently available default architectures include:

\begin{itemize}
    \item \textbf{MLP} - This is a standard Multilayer Perceptron (MLP) which flattens the input into a single vector of length $\#$input time delays $\times$ $\#$input components and passes the features through a set of fully connected layers to produce the final output. 
    % For more detail on this architecture see the works of Goodfellow et al.~\cite{Goodfellow-et-al-2016} or Prince~\cite{prince2023understanding}.
    This architecture applies no time series related inductive bias, making it a simple baseline to determine expected model performance without considering the temporal nature of the data.
    \item \textbf{TCN} - The Temporal Convolutional Network (TCN) includes a convolutional stage that performs one dimensional convolutions across the time domain. Additionally, it uses dilated causal convolutions and padding to ensure that there is no information leakage from future values. It is based on the work by Bai et al.~\cite{bai2018empirical} and is a fitting choice if the model should capture relatively short-term and high-frequency relationships in time~\cite{wang2024tssurvey}.
    \item \textbf{CNN} - The Convolutional Neural Network (CNN) is a similar model to the TCN, in that it performs one dimensional convolutions across the time domain. It is made non-causal by removing padding and clipping. In scenarios where causality is not required, it could be a good alternative to the TCN.
    \item \textbf{LSTM} - The Long Short-Term Memory (LSTM) architecture uses recurrent connections to update a hidden state and an internal/cell state sequentially, as the model passes over the time domain. For more information about LSTMs, see Hochreiter and Schmidhuber~\cite{10.1162/neco.1997.9.8.1735} and Zhang et al.~\cite{zhang2023dive}.
    \item \textbf{LSTMv2} - This is an alternative version of the LSTM architecture with additional capabilities for mitigating unstable gradients (e.g. layer normalization and residual connections). It can also be made stateful,  meaning this architecture could be a good choice if longer-term dependencies across time are expected.
    \item \textbf{TFT} - This is a Temporal Fusion Transformer-style architecture. It omits the encoding of exogenous inputs, since they are not currently supported within \textit{KnowIt}, but makes use of variable selection, gating mechanisms, LSTM-based encoding, and Interpretable Multi-Head Attention. It is based on the work by Lim et al.~\cite{lim2021temporal} and wraps the \textit{LSTMv2} module for its LSTM encoding stage. This is a good choice of architecture for tasks that require the modeling of long-term dependencies provided sufficient data and compute.
\end{itemize}

% As mentioned in Section~\ref{sec: problem class}, all current \textit{KnowIt} default architectures (including the LSTM) are stateless. This means that each input sequence is considered in isolation and hidden states are not maintained across different sequences or batches. Correspondingly, only time dependencies within each sequence is considered for parameter updates during training. We are currently working towards introducing optional statefullness and adding a transformer based model, namely a temporal fusion transformer~\cite{lim2021temporal}.

\subsection{Interpretability techniques}
\label{sec: interpret}

% There are many approaches to interpret a DNN's decisions. Different approaches consider different characteristics; e.g. weights, circuits, gradients, data selectivity, or inherently interpretable structure. In order to enforce the overall theme of \textit{extensibility}, \textit{KnowIt} uses a modular interpretability framework, where a user can easily define a new class of interpreter that inherets from a common module within \textit{KnowIt}.

Current interpretability techniques in \textit{KnowIt} build on the concept of {\it feature attribution}~\cite{lundberg2017unified}: the extent to which the value of a given feature affects a specific prediction. 
These values can be combined in different ways in order to analyze {\it feature interactions}~\cite{herbinger2024decomposing}, that is, the effect the value of one feature has on the attribution of another. 
By combining the absolute values of attributions over multiple predictions, feature values and/or timesteps, it is possible to obtain a better understanding of {\it feature importance} in specific scenarios.  

\textit{KnowIt} provides seamless integration with feature attribution techniques provided by {\it Captum}~\cite{kokhlikyan2020captum}.
% \todo{To discuss: how much detail about techniques, or just mention and reference? Or even less?}
%
Currently implemented attributers include DeepLift\footnote{\url{https://captum.ai/api/deep_lift.html}}, DeepLiftShap\footnote{\url{https://captum.ai/api/deep_lift_shap.html\#captum.attr.DeepLiftShap}}, and Integrated Gradients\footnote{\url{https://captum.ai/api/integrated_gradients.html}}.
Once feature attributions have been extracted, these can be visualized and analyzed.
% \todo{How do you feel about a pic here? Would also like to touch on ref values, scaling, selection,...}
This approach provides a basis for the development of additional visualization capabilities, an ongoing development effort. 

\section{Code design}
\label{sec: code design}

\begin{figure}[h]
    \centering
    \includegraphics[width=0.75\linewidth]{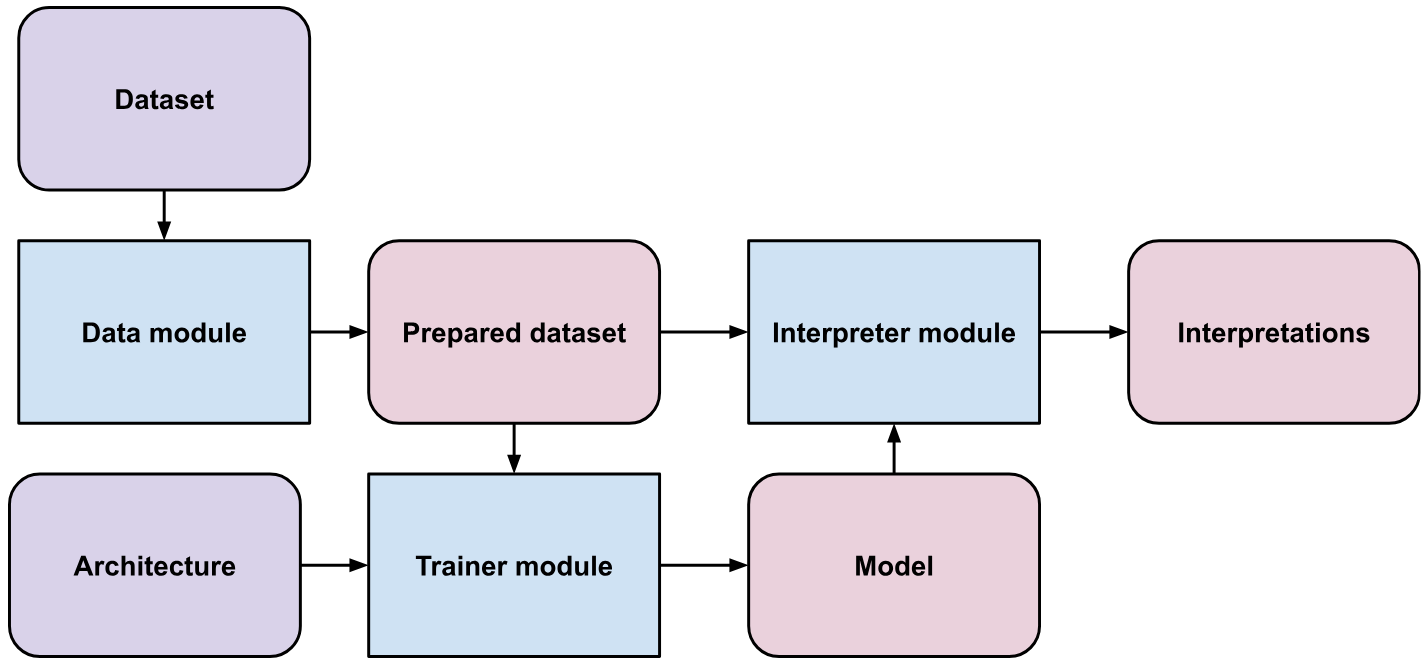}
    \caption{General module configuration in \textit{KnowIt}. Blue elements represent code modules, purple elements represent static data structures, and pink elements represent data structures that are dynamically generated based on what is being processed.}
    \label{fig:modules}
\end{figure}

\textit{KnowIt} uses a modular design, consisting of three main modules and an overarching ``KnowIt'' module with which the user guides operations. Fig.~\ref{fig:modules} shows a flow chart illustrating the general interaction between the three main modules once a dataset and deep learning architecture has been imported or defaults are selected. See Fig.~\ref{fig:class structure} in Appendix\ref{app sec: class structure} for a more detailed overview of these modules and the classes that they are made up of.

The data module receives the path to an imported dataset on disk and prepares it for training or interpreting. The trainer module receives a deep learning architecture class definition and the prepared dataset. It produces a model of the data using the given architecture. The interpreter module receives the prepared dataset and trained model and produces an interpretation of the model with respect to the data. Note that user arguments are also provided to define details related to each of these steps.

\textbf{Data module}\\
The data module facilitates most aspects of time series data within the \textit{KnowIt} framework. It imports new raw data and prepares imported data for training or interpreting. This includes operations like the sampling, splitting, and scaling of relevant parts of the data in order to be processed by models. All data is imported as a \textit{pandas}~\cite{pandas} dataframe, but it can be imported from multiple file types (e.g. pickle, csv, parquet, json). The imported data is kept on disk as a partitioned parquet file with a pickle file containing meta data.

\textbf{Trainer module}\\
The trainer module facilitates the training of new models and evaluation of trained models through the \textit{PyTorch Lightning}~\cite{Falcon_PyTorch_Lightning_2019} framework. It expects a set of \textit{PyTorch}~\cite{Ansel_PyTorch_2_Faster_2024} dataloaders and a \textit{PyTorch} compatible architecture class.

\textbf{Interpreter module}\\
The interpreter module facilitates the interpretation of trained models using the \textit{Captum}~\cite{kokhlikyan2020captum} framework. This is currently limited to feature attribution methods, however the module is purposely built to be expanded to different interpretation paradigms.

\section{Operational flow}
\label{sec: operattional flow}

The user interacts with \textit{KnowIt} by importing the main ``KnowIt'' module, instantiating an object of it and then sending it the required key word arguments for each kind of operation. In Section~\ref{sec: use case} we describe the general use case for \textit{KnowIt} and in Section~\ref{sec: use example} we provide links to online tutorials illustrating concrete examples of using the toolkit.

\subsection{General use case}
\label{sec: use case}

% Explain the general use case and expected steps (get data, prep data, import data, train prelim models and explore, set up HP sweep, get good model, interpret, optional Tune further or tweak). Use a nice graph. How does the output structure look? 

\begin{figure}[b]
    \centering
    \includegraphics[width=0.6\linewidth]{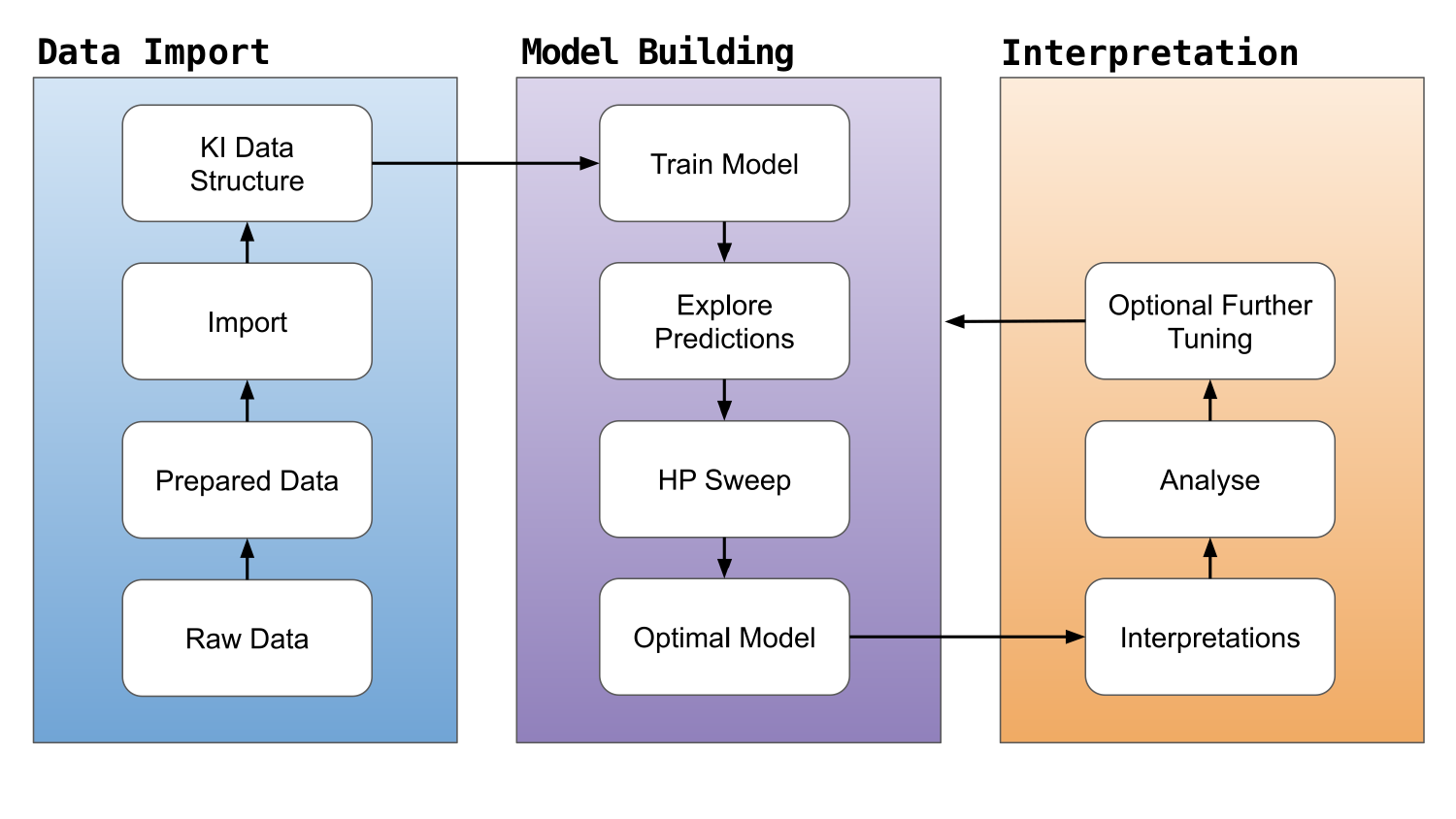}
    \caption{The expected steps illustrating a complete cycle through \textit{KnowIt}.}
    \label{fig:expected_steps}
\end{figure}

Given a raw time series dataset, the general use case of \textit{KnowIt} is to extract knowledge from the time series dataset using a deep learning model and interpretability techniques. This is called \emph{knowledge discovery}. In order to perform knowledge discovery using \emph{KnowIt}, a series of steps need to be performed. The expected steps for \textit{KnowIt} are shown in Figure~\ref{fig:expected_steps} and are divided into three broad categories, namely, data import, model building, and interpretation. 

\paragraph{Data Import} Starting with a raw time series dataset, the user needs to first prepare the data prior to importing it into \textit{KnowIt}. Typically, this involves some light data processing and the creation of metadata. Once the data is prepared, the user can import the data into \textit{KnowIt}, which will format the data into a structure that \textit{KnowIt} understands. This data structure is saved in the user's experiment output directory under the ``custom\_datasets'' subfolder.

\paragraph{Model Building} Once the data is imported, \textit{KnowIt} is ready for model training. In order to extract reliable knowledge from the dataset, a model with appropriate hyperparameters needs to be selected. There are two options in this case:
\begin{itemize}
    \item Empirical-tuning: A user trains a single model at a time and tunes the model hyperparameters by hand according to some model selection criteria. This is not recommended for data with which the user is not familiar.
    \item Hyperparameter sweep: A user trains multiple models through a hyperparameter sweep and selects the best model according to some model selection criteria.
\end{itemize}
In the case of empirical-tuning, \textit{KnowIt} will save a model checkpoint, as well as various metrics, in the user's experiment output directory. By providing a visualizer flag, a user can optionally view the model training curves in the same output directory. In contrast, in the case of hyperparameter sweeps, \textit{KnowIt} can ``consolidate'' the sweep by saving the best model, or the user can choose to save no models. Sweep related model training metrics are logged to \textit{Weights and Biases}~\cite{wandb} during hyperparameter sweeps.

Finally, in addition to the quantitative metrics provided during training, a user also has the option to visually inspect the model's fit to the training, validation, or evaluation set.

\paragraph{Interpretation} Once the user has built a model with a good fit to the data, \textit{KnowIt} can be used to extract interpretations. Given some user argument (e.g. to interpret random prediction points or prediction points where the model performs best), \textit{KnowIt} uses feature attribution techniques to assign an importance score to each input feature relative to an output component at any given point in time. % Currently, \textit{KnowIt} is packaged with three attribution techniques, namely, Integrated Gradients, DeepLift, and DeepLiftSHAP. However, due to the modularity of \textit{KnowIt}, further attribution techniques can be easily added. The attributions are stored in the user's output directory. Optionally, a user can also view a heatmap representation of the attributions.

\paragraph{Experiment output Structure} All \textit{KnowIt} outputs are saved in a custom ``experiment output directory''. This directory is intuitively structured as follows:
\begin{itemize}
    \item Custom datasets: stores a user's imported data in a structure created by \textit{KnowIt}.
    \item Models: stores all model checkpoints, metrics, sweeps, and interpretations.
    \item Custom architectures: stores any custom architecture definitions that a user defines.
\end{itemize}
\textit{KnowIt} stores and manages all outputs in this directory in a specific format. See the \textit{Experiment Structure}~\footnote{\url{https://must-deep-learning.github.io/KnowIt/markdowns/guides/result_structure_readme.html}}  guide in the documentation for more details on this format.

% Each of these parent folders will contain a subfolder that is named as the user's experiment name (defined in a user's experiment script). All relevant files are stored in these respective subfolders.

% \section{Datasets, models, tasks (TT)}

% Explain what forms time series is expected to be, how it is stored, split and sampled.

% Elaborate on how to define various types of time series modeling tasks.

% Explain how each of the current architectures function. Define the inputs and output formats, which are common to all of them.

% \section{Interpretability (MD)}
% Explain what feature attributions, DeepliftShap, DeepShap, and integrated gradients are, and what they usually look like for time series data.

% \section{Lessons learned (TT)}
% Mention some of the quirks of working with deep time series models and time series data.

\subsection{Usage example}
\label{sec: use example}
% % A very brief illustration (with code) of going from importing data to interpretation. Use as little code as possible.
% In this section, we provide a minimalist example that illustrates the steps in Section~\ref{sec: operattional flow}.

% \begin{figure}
%     \centering
%     \includegraphics[trim=3cm 3cm 3cm 6cm, clip, width=0.5\linewidth]{figures/code-snapshot.png}
%     \caption{Caption}
%     \label{fig:enter-label}
% \end{figure}
% \todo{Which looks better - the snapshot or the custome code text?}
% Or:
% \begin{lstlisting}[language=Python, caption=KnowIt usage example]
% from knowit import KnowIt

% # instantiate KnowIt and create a new output directory
% ki = KnowIt(
%     custom_exp_dir="/home/user/path/to/output/directory",
% )

% # import a prepared time series data set
% ki.import_dataset(
%     kwargs={
%         "data_import": "/home/user/path/to/dataset.pickle",
%     },
% )

% # train a model
% model_name = "my_model"
% ki.train_model(
%     model_name=model_name,
%     kwargs={
%         "data": {"data_args..."},
%         "arch": {"arch_args..."},
%         "trainer": {"trainer_args..."},
%     },
% )

% # extract interpretations
% ki.interpret_model(
%     model_name=model_name,
%     kwargs={
%         "interpreter": {"interpret_args..."},
%     },
% )
% \end{lstlisting}

For an example of the steps described in the previous subsection we point the reader to the following two tutorials:
\begin{itemize}
    \item \textbf{Basics}~\footnote{\url{https://must-deep-learning.github.io/KnowIt/markdowns/tutorials/basics/basics_tut_readme.html}}: This tutorial shows how to load, prepare, and import a new dataset into \textit{KnowIt}, as well as train, evaluate, and interpret a new model of this dataset.
    \item \textbf{Basic sweep}~\footnote{\url{https://must-deep-learning.github.io/KnowIt/markdowns/tutorials/basic_sweep/basic_sweep_tut_readme.html}}: This tutorial shows how to perform a basic hyperparameter sweep in the \textit{KnowIt} framework.
    \item \textbf{Statefulness}~\footnote{\url{https://must-deep-learning.github.io/KnowIt/markdowns/tutorials/stateful/stateful_tut_readme.html}}: This tutorial illustrates the basic idea behind stateful training.
\end{itemize}

\section{Related toolkits}
\label{sec: related toolkits}

% See full comparison here:
% https://docs.google.com/spreadsheets/d/1zEpQffYsUVq1EfGMgpR0K8iNoYjQu8pvRU4p6DJKZWI/edit?gid=148002202#gid=148002202

\textit{KnowIt} is not the first toolkit focusing on time series modeling with machine learning methods. See Table.~\ref{tab: related} for reference,
where we compare \textit{KnowIt} with popular open source Python packages focused on time series modeling. 
All toolkits support the modeling of given multivariate time series data, provide extensive support for deep learning models\footnote{Either as the core modeling approach (Primary) or in conjunction with classic machine learning methods (Dual)}, have hyperparameter tuning support, and have been updated during 2024 at the latest. 
We note three main differences between \textit{KnowIt} and alternatives.

First, we find that no alternative has the interpretability of the resulting models as a main focus. 
Several toolkits include ``intrinsically interpretable'' models~\cite{tsai, JMLR:v23:21-1177, olivares2022library_neuralforecast, pytorch_forcasting, gluonts_jmlr} 
and some provide limited explainability support in the form of feature attributions based on Shapley values~\cite{JMLR:v23:21-1177, godfried2020flowdb}. 
In contrast, \textit{KnowIt} %has a primary focus on interpretability with care taken 
was designed to ensure that all models can be interpreted with a growing list of available interpretability techniques.

Secondly, alternative toolkits typically boast an extensive list of deep learning architectures to choose from but provide  limited support for importing custom architectures. 
Some provide guides for contributors who want to add to the relevant toolkit's library of architectures~\cite{du2023pypots, bhatnagar2021merlion, olivares2022library_neuralforecast}, 
and others provide user guides on how to import custom architectures~\cite{pytorch_forcasting, gluonts_jmlr}. 
%This is typically presented as a supplementary feature. 
In \textit{KnowIt}, importing custom architectures is designed as part of the general use case, with well-defined criteria (see \textit{Architecture How-to}~\footnote{\url{https://must-deep-learning.github.io/KnowIt/markdowns/guides/archs_readme.html}} documentation) to ensure that any model acts as a drop-in replacement for the existing collection of default architectures.

Finally, all alternatives heavily emphasize forecasting in their collection of supported tasks, with the majority~\cite{olivares2022library_neuralforecast, godfried2020flowdb, pytorch_forcasting, gluonts_jmlr, JMLR:v23:21-1177, bhatnagar2021merlion, autots} presenting forecasting as the main focus and other tasks (such as classification or anomaly detection) a supplementary function, if included at all. 
In the \textit{KnowIt} toolkit we do not focus on modeling time series from the perspective of forecasting, explicitly. Rather, as described in Section~\ref{sec: problem class}, we develop a general regression and classification framework, from which other tasks can be defined directly or indirectly, resulting in a flexible and powerful modeling environment.

\begin{table}[H]
\caption{Comparison of \textit{KnowIt} with alternative time series modeling toolkits. \textit{FORE} - Forecasting, \textit{CLASS} - Classification, \textit{ANOD} - Anomaly detection, \textit{REG} - Regression, \textit{IMPU} - Imputation, \textit{CLUS} - Clustering, \checkmark - directly supported, $\ast$ - indirectly supported.}
\label{tab: related}
\begin{tabular}{cccccccccc}
\textbf{Toolkit}                                                                     & \textbf{\begin{tabular}[c]{@{}c@{}}DL\\ Focus\end{tabular}} & \textbf{\begin{tabular}[c]{@{}c@{}}Interpret-\\ ability\end{tabular}} & \textbf{\begin{tabular}[c]{@{}c@{}}Custom \\ models\end{tabular}}                  & \textbf{FORE}                   & \textbf{CLASS}                  & \textbf{ANOD}                   & \textbf{REG}                    & \textbf{IMPU}                   & \textbf{CLUS}                   \\ \hline
\multicolumn{1}{|c|}{KnowIt (Ours)}                                                         & \multicolumn{1}{c|}{Primary}                                & \multicolumn{1}{c|}{Primary}                                          & \multicolumn{1}{c|}{Core use}                                                     & \multicolumn{1}{c|}{$\ast$}          & \multicolumn{1}{c|}{\textbf{\checkmark}} & \multicolumn{1}{c|}{$\ast$}          & \multicolumn{1}{c|}{\textbf{\checkmark}} & \multicolumn{1}{c|}{$\ast$}          & \multicolumn{1}{c|}{}           \\ \hline
\multicolumn{1}{|c|}{Neuralforecast~\cite{olivares2022library_neuralforecast}}                                                 & \multicolumn{1}{c|}{Primary}                                & \multicolumn{1}{c|}{Secondary}                                        & \multicolumn{1}{c|}{\begin{tabular}[c]{@{}c@{}}Contributor \\ guide\end{tabular}} & \multicolumn{1}{c|}{\textbf{\checkmark}} & \multicolumn{1}{c|}{$\ast$}          & \multicolumn{1}{c|}{}           & \multicolumn{1}{c|}{}           & \multicolumn{1}{c|}{}           & \multicolumn{1}{c|}{}           \\ \hline
\multicolumn{1}{|c|}{tsai~\cite{tsai}}                                                           & \multicolumn{1}{c|}{Primary}                                & \multicolumn{1}{c|}{Secondary}                                        & \multicolumn{1}{c|}{None}                                                         & \multicolumn{1}{c|}{\textbf{\checkmark}} & \multicolumn{1}{c|}{\textbf{\checkmark}} & \multicolumn{1}{c|}{}           & \multicolumn{1}{c|}{\textbf{\checkmark}} & \multicolumn{1}{c|}{\textbf{\checkmark}} & \multicolumn{1}{c|}{}           \\ \hline
\multicolumn{1}{|c|}{\begin{tabular}[c]{@{}c@{}}Flow \\ Forecast~\cite{godfried2020flowdb}\end{tabular}}       & \multicolumn{1}{c|}{Primary}                                & \multicolumn{1}{c|}{Secondary}                                        & \multicolumn{1}{c|}{None}                                                         & \multicolumn{1}{c|}{\textbf{\checkmark}} & \multicolumn{1}{c|}{$\ast$}          & \multicolumn{1}{c|}{$\ast$}          & \multicolumn{1}{c|}{}           & \multicolumn{1}{c|}{}           & \multicolumn{1}{c|}{}           \\ \hline
\multicolumn{1}{|c|}{\begin{tabular}[c]{@{}c@{}}PyTorch \\ Forecasting~\cite{pytorch_forcasting}\end{tabular}} & \multicolumn{1}{c|}{Primary}                                & \multicolumn{1}{c|}{None}                                             & \multicolumn{1}{c|}{\begin{tabular}[c]{@{}c@{}}User \\ guide\end{tabular}}        & \multicolumn{1}{c|}{\textbf{\checkmark}} & \multicolumn{1}{c|}{}           & \multicolumn{1}{c|}{}           & \multicolumn{1}{c|}{}           & \multicolumn{1}{c|}{}           & \multicolumn{1}{c|}{}           \\ \hline
\multicolumn{1}{|c|}{GluonTS~\cite{gluonts_jmlr}}                                                        & \multicolumn{1}{c|}{Primary}                                & \multicolumn{1}{c|}{None}                                             & \multicolumn{1}{c|}{\begin{tabular}[c]{@{}c@{}}User \\ guide\end{tabular}}        & \multicolumn{1}{c|}{\textbf{\checkmark}} & \multicolumn{1}{c|}{}           & \multicolumn{1}{c|}{}           & \multicolumn{1}{c|}{}           & \multicolumn{1}{c|}{}           & \multicolumn{1}{c|}{}           \\ \hline
\multicolumn{1}{|c|}{Darts~\cite{JMLR:v23:21-1177}}                                                          & \multicolumn{1}{c|}{Dual}                                   & \multicolumn{1}{c|}{Secondary}                                        & \multicolumn{1}{c|}{None}                                                         & \multicolumn{1}{c|}{\textbf{\checkmark}} & \multicolumn{1}{c|}{}           & \multicolumn{1}{c|}{\textbf{\checkmark}} & \multicolumn{1}{c|}{}           & \multicolumn{1}{c|}{}           & \multicolumn{1}{c|}{}           \\ \hline
\multicolumn{1}{|c|}{PyPOTS~\cite{du2023pypots}}                                                         & \multicolumn{1}{c|}{Dual}                                   & \multicolumn{1}{c|}{None}                                             & \multicolumn{1}{c|}{\begin{tabular}[c]{@{}c@{}}Contributor \\ guide\end{tabular}} & \multicolumn{1}{c|}{\textbf{\checkmark}} & \multicolumn{1}{c|}{\textbf{\checkmark}} & \multicolumn{1}{c|}{\textbf{\checkmark}} & \multicolumn{1}{c|}{}           & \multicolumn{1}{c|}{\textbf{\checkmark}} & \multicolumn{1}{c|}{\textbf{\checkmark}} \\ \hline
\multicolumn{1}{|c|}{Merlion~\cite{bhatnagar2021merlion}}                                                        & \multicolumn{1}{c|}{Dual}                                   & \multicolumn{1}{c|}{None}                                             & \multicolumn{1}{c|}{\begin{tabular}[c]{@{}c@{}}Contributor \\ guide\end{tabular}} & \multicolumn{1}{c|}{\textbf{\checkmark}} & \multicolumn{1}{c|}{}           & \multicolumn{1}{c|}{\textbf{\checkmark}} & \multicolumn{1}{c|}{}           & \multicolumn{1}{c|}{}           & \multicolumn{1}{c|}{}           \\ \hline
\multicolumn{1}{|c|}{AutoTS~\cite{autots}}                                                         & \multicolumn{1}{c|}{Dual}                                   & \multicolumn{1}{c|}{None}                                             & \multicolumn{1}{c|}{None}                                                         & \multicolumn{1}{c|}{\textbf{\checkmark}} & \multicolumn{1}{c|}{}           & \multicolumn{1}{c|}{}           & \multicolumn{1}{c|}{}           & \multicolumn{1}{c|}{}           & \multicolumn{1}{c|}{}           \\ \hline
\end{tabular}
\end{table}

\section{Conclusion}
% \todo{Do right at end}

We introduce the {\it KnowIt} toolkit: a flexible environment for modeling and analyzing a broad range of time series tasks. 
The main assumption underlying framework design is that the data being modeled contain features that are equidistant (or rendered to be equidistant). 
The framework is designed with few other assumptions about the nature of the tasks, resulting in a highly customisable and easy-to-use environment. 
As model choices are separated from interpretability choices through clearly specified interfaces, it  is straightforward to add either a new architecture or analysis technique, which then becomes available throughout the framework.

Our goal is to continue to maintain and expand \textit{KnowIt} to foster application, collaboration, and investigation of deep time series modeling and interpretation, an important and relatively under-investigated field of study. We welcome feedback through our issue list on GitHub\footnote{\url{https://github.com/MUST-Deep-Learning/KnowIt/issues}}.

% Mention current applications (penguin tracking? space weather? EMG?). 
% Next: Expand toolkit (Harmen, Leah, Almaro). Maintain toolkit. Foster collaboration. Broader impact etc.

% \section{Acknowledgements}
% This project is made possible due to funding and support from: MUST Deep Learning, North-West University (NWU), South Africa, National Institute for Theoretical \& Computational Sciences (NITheCS), South Africa, and Centre for Artificial Intelligence Research (CAIR), South Africa.

\section{Acknowledgments and Disclosure of Funding}

This work is based on research supported in part by the National Research Foundation of South Africa (Ref Numbers \textit{PSTD23042898868}, \textit{RA211019646111}).

% \section{Other todo's}
% \begin{itemize}
%     % \item Confirm US vs UK spelling and auto-check
%     % \item Sentence about name and final goal (KD).
%     % \item Probabilistic forecasting?
%     % \item Journal: JMLR? If githubbed... perhaps? [If not, what is second choice?]
%     % \item Does JMLR allow preprint arXiving? If so, release there first.
%     \item Spell/grammar check documentation
% \end{itemize}

% "Since we specifically want to honor the effort of turning a method into a highly usable piece of software, prior publication of the method is admissible, as long as the software has not been published elsewhere. If the software has been the main content of a paper appearing in a peer reviewed conference or journal, then there should be a document in the code repository (referred to in the cover letter of the submission), listing the software package's improvements and extensions. It is hoped that the open source track will motivate the machine learning community towards open science, where open access publishing, open data standards and open source software foster research progress. " \url{https://www.jmlr.org/mloss/mloss-info.html}

 \bibliographystyle{splncs04}
 % \clearpage
 \bibliography{references}

 \appendix

 \section{Problem class}
\label{app sec: problem class}

In supervised machine learning, a dataset consists of input-output pairs $(x, y)$, where $y = f(x) + \epsilon$ and the task is to model the unknown function $f$. In this section, we provide details concerning how a time series dataset is transformed to a supervised machine learning task for \emph{KnowIt}.

Let $\{x_t \}_{t=1}^{n}$ denote an ordered sequence of values (i.e. a time series), where $x_t \in \mathbb{R}^{c_{in}}$. Similarly, let $\{y_t\}_{t=1}^{n}$, where $y_t \in \mathbb{R}^{c_{out}}$. We assume that there is some unknown relationship between $\{x_t \}_{t=1}^{n}$ and $\{y_t \}_{t=1}^{n}$; more precisely, given some ordered subset $\mathcal{X} \subset \{x_t \}_{t=1}^{n}$ and $\mathcal{Y} \subset \{y_t \}_{t=1}^{n}$, there exists an unknown function $f$ such that
\begin{equation}
    \mathcal{Y} = f(\mathcal{X}).
\end{equation}

To formulate the time series task as a supervised machine learning task, we construct the following ordered subsets: let $a,b \in \mathbb{Z}$, $a \leq b$, and $t\in [n]$. Then, we define \footnote{Note that we denote our interval in somewhat unconventional notation as  $[t+a,..., t,..., t+b]$. This is because \emph{KnowIt} offers flexibility for the choice of integers $a,b$. For example, $a=-2, b=5$ gives the interval $[t-2,...,t,...,t+5]$; in contrast, $a=2,b=5$ gives $[t+2,...,t+5]$. The prediction point $t$ need not be an element of the interval.}
\begin{equation}
    \mathcal{X}_{t, a, b} := \{x_i : i \in [t+a,...,t,...,t+b] \} \subset \{x_t \}_{t=1}^{n}.
\end{equation}
Similarly, for $c,d \in \mathbb{Z}$ and $c \leq d$, we define
\begin{equation}
    \mathcal{Y}_{t, c, d} := \{y_i : i \in [t+c,...,t,...,t+d] \} \subset \{y_t \}_{t=1}^{n}.
\end{equation}
Note that for values of $t$ where any $i < 1$ or $i > n$, padding is used or these sets are dropped from consideration. Note also that the choice of integers $a,b,c,d$ are hyperparameters to be determined. We refer to each $t$ as a \emph{prediction point}. The ordered collection of these sets $\{ (\mathcal{X}_{t,a,b}, \mathcal{Y}_{t,c,d}) \}_{t=1}^{n}$ forms our input-output pairs.

Finally, since deep learning models operate on vectors or matrices, we construct the following matrices: for a $\mathcal{X}_{t,a,b} \in \{ \mathcal{X}_{t,a,b}\}_{t=1}^{n}$ and $\mathcal{Y}_{t,c,d} \in \{ \mathcal{Y}_{t,c,d}\}_{t=1}^{n}$, we define the matrices
\begin{equation}
    X_{t,a,b} = \begin{bmatrix}
        x_{t+a}^T \\
        \vdots \\
        x_{t}^T \\
        \vdots \\
        x_{t+b}^T
    \end{bmatrix} \in \mathbb{R}^{(|a| + |b| + 1)\times c_{in}}, \quad 
    Y_{t,c,d} = \begin{bmatrix}
        y_{t+c}^T \\
        \vdots \\
        y_{t}^T \\
        \vdots \\
        y_{t+d}^T
    \end{bmatrix} \in \mathbb{R}^{(|c| + |d| + 1)\times c_{out}}.
\end{equation}
Note that each row in $X_{t,a,b}$ corresponds to all input components at a single time delay defined relative to the prediction point $t$. Correspondingly, each column in $X_{t,a,b}$ embodies a single input component across multiple time delays defined relative to the prediction point $t$. The arrangement of $Y_{t,c,d}$ follows the same structure. We, therefore, have the final dataset $\{ (X_{t,a,b}, Y_{t,c,d})\}_{t=1}^{n}$ which is ready for a deep learning model.

This approach to time series modeling might seem convoluted, however, it provides a lot of flexibility in the type of tasks that we can define. For example,
\begin{itemize}
    \item If the components in $Y_{t,c,d}$ are also contained in $X_{t, a,b}$ and $a\leq b < c \leq d$, we are performing an autoregressive forecasting task.
    \item If $c=d$, we are making a single-step prediction.
    \item If $c_{out}=1$, we are performing a univariate task as opposed to a multivariate task.
    \item If $a \leq b < c \leq d$, our model is respecting temporal causality (local to the prediction).
    \item For time series classification, if $y_t \in \{0,...,K-1 \}$, we can use one-hot encoding to form the vector $y_t \in \mathbb{R}^{K}$. We can then proceed in the same manner as above.
\end{itemize}

 \section{Class interaction}
\label{app sec: class structure}

\begin{figure}[h]
    \centering    \includegraphics[width=\linewidth]{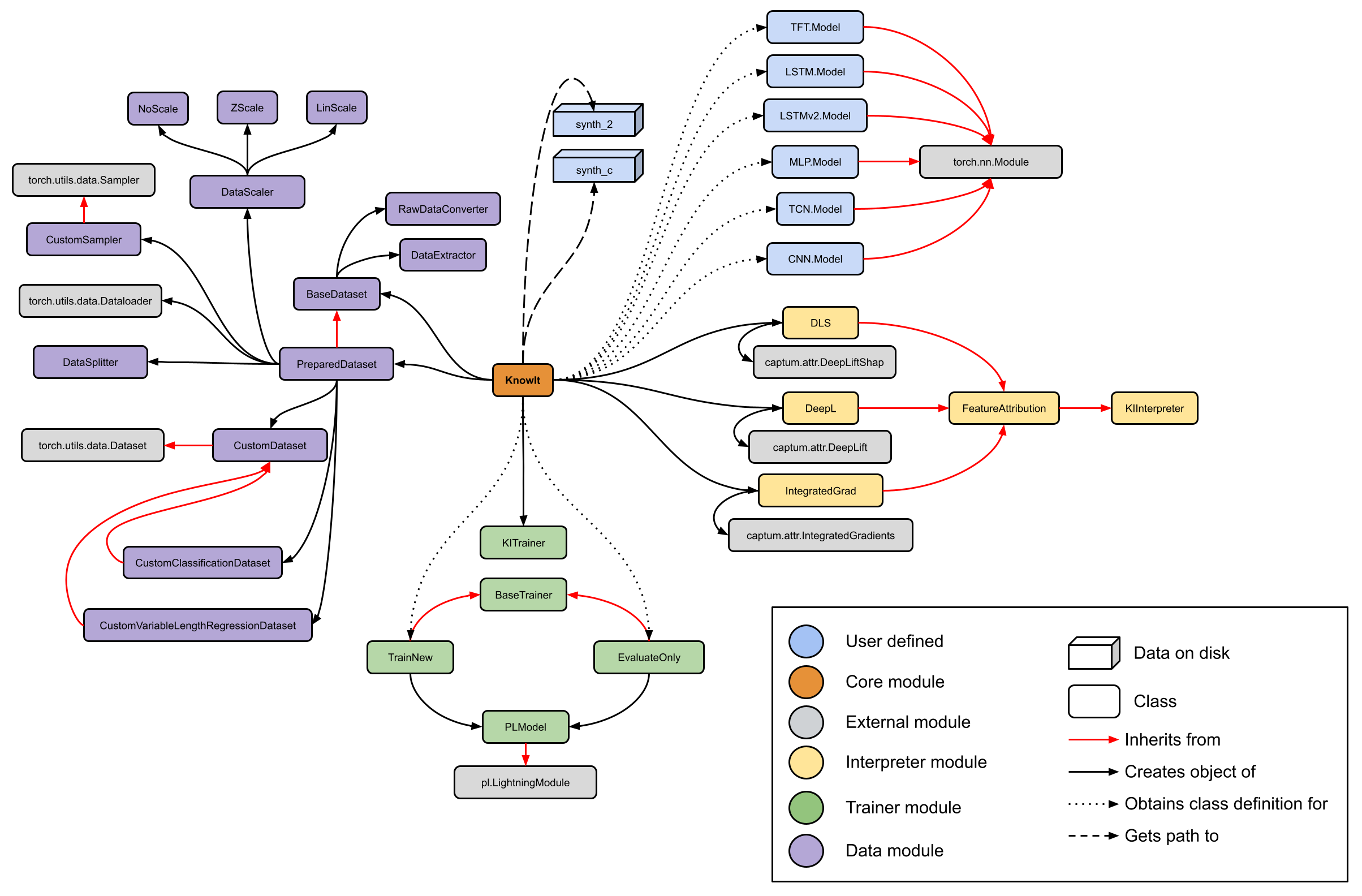}
    \caption{High level class interaction in \textit{KnowIt}.}
    \label{fig:class structure}
\end{figure}

Fig.~\ref{fig:class structure} provides a high level overview of all the classes in the \textit{KnowIt} framework. The user always interacts with the core ``KnowIt'' class, which makes use of the relevant peripheral classes.

\end{document}